\documentclass{article}

\usepackage{PRIMEarxiv}

\usepackage[utf8]{inputenc} 
\usepackage[T1]{fontenc}    
\usepackage{hyperref}       
\usepackage{url}            
\usepackage{booktabs}       
\usepackage{amsfonts}       
\usepackage{nicefrac}       
\usepackage{microtype}      
\usepackage{lipsum}
\usepackage{fancyhdr}       
\usepackage{graphicx}       
\graphicspath{{media/}}     
\usepackage{verbatim}
\usepackage{amssymb}
\usepackage{import}
\usepackage[natbibapa]{apacite}
\usepackage{amsmath} 
\usepackage{multirow} 
\usepackage{setspace}
\usepackage[switch]{lineno}
\usepackage{float}
\usepackage{newtxmath}
\usepackage{xcolor}

\title{MAME: Multidimensional Adaptive Metamer Exploration with \\
Human Perceptual Feedback
}

\author{
  Mina Kamao\\
  Graduate School of Information Science and Technology\\
  The University of Tokyo, Tokyo, Japan\\
  \texttt{kmmina0902@g.ecc.u-tokyo.ac.jp}\\
  \And
  Hayato Ono \\
  Graduate School of Information Science and Technology\\
  The University of Tokyo, Tokyo, Japan\\
  \texttt{hayato-0628@g.ecc.u-tokyo.ac.jp}\\
  \And
  Ayumu Yamashita \\
  1. Graduate School of System Informatics\\
  Kobe University, Hyogo, Japan\\
  2. Graduate School of Information Science and Technology\\
  The University of Tokyo, Tokyo, Japan\\
  \texttt{ayumu722@people.kobe-u.ac.jp}\\
  \And
  Kaoru Amano \\
  Graduate School of Information Science and Technology\\
  The University of Tokyo, Tokyo, Japan\\
  \texttt{kaoru\_amano@ipc.i.u-tokyo.ac.jp}\\
  \And
  Masataka Sawayama \thanks{Corresponding authors: Masataka Sawayama and Mina Kamao}\\
  1. Graduate School of Information Science and Technology\\
  Hokkaido University, Hokkaido, Japan\\
  2. Prometech CG Research, Tokyo, Japan\\
  3. Graduate School of Information Science and Technology\\
  The University of Tokyo, Tokyo, Japan\\
  \texttt{masataka\_sawayama@ist.hokudai.ac.jp}\\
}

\begin{document}

\onehalfspacing

\maketitle
\thispagestyle{fancy}


\begin{abstract}
Alignment between human brain networks and artificial models has become an active research area in vision science and machine learning. A widely adopted approach is identifying "metamers," stimuli physically different yet perceptually equivalent within a system. However, conventional methods lack a direct approach to searching for the human metameric space. Instead, researchers first develop biologically inspired models and then infer about human metamers indirectly by testing whether model metamers also appear as metamers to humans. Here, we propose the Multidimensional Adaptive Metamer Exploration (MAME) framework, enabling direct, high-dimensional exploration of human metameric spaces through online image generation guided by human perceptual feedback. MAME modulates reference images across multiple dimensions based on hierarchical neural network responses, adaptively updating generation parameters according to participants' perceptual discriminability. Using MAME, we successfully measured multidimensional human metameric spaces within a single psychophysical experiment. Experimental results using a biologically plausible CNN model showed that human discrimination sensitivity was lower for metameric images based on Gram-matrix representations derived from low-level CNN features than for those derived from high-level CNN features. The finding suggests a relatively worse alignment between the metameric spaces of humans and the CNN model for low-level processing compared to high-level processing. Counterintuitively, given recent discussions on alignment at higher representational levels, our results highlight the importance of early visual computations in shaping biologically plausible models. Our MAME framework can serve as a future scientific tool for directly investigating the functional organization of human vision.

\end{abstract}

\begin{quote}
\end{quote}

\section{Introduction}
\label{introduction}
Recent advancements in artificial neural networks (ANNs) have not only improved their ability to solve visual tasks but have also positioned them as valuable analogies for understanding human visual processes. In particular, deep convolutional neural networks (CNNs) have been noted for their structural and functional similarities to the hierarchical processing observed in biological vision. These models have demonstrated considerable success in approximating the information flow from the primary visual cortex to higher visual areas in the ventral visual stream \citep{yamins2016using}. However, numerous studies have reported significant divergences between the behavioral responses of humans and ANN models \citep{geirhos2018imagenet, geirhos2018generalisation, szegedy2013intriguing}. Consequently, extensive research efforts have been directed toward achieving a functional alignment between human vision and artificial models \citep{geirhos2021partial}.

\subsection{Metamer exploration as a tool for functional understanding of cognitive neural processes}

One widely adopted approach for functionally aligning models with humans is the identification of "metamers." Metamers refer to sets of input stimuli that are physically different yet processed as equivalent by a given system, such as humans or models. Here, consider an unknown system whose functional properties we aim to uncover. If the system produces the same output for different sets of physical inputs, i.e., metamers, it implies the existence of common factors to which the system is sensitive. Identifying these common factors would provide insight into what the system functionally computes.

Through this metamer-based approach, various visual functions have been uncovered in the history of vision sciences. A classic example is the finding that human color vision is mediated by three types of sensors (reviewed by \citep{gegenfurtner2003cortical}). The discovery that humans do not directly sense light as a function of wavelength was based on experimental results showing that physically different spectral compositions of light can appear identical to human observers, i.e., they form metamers.

Furthermore, this approach has also been used to investigate hierarchical information processing in the brain. In the context of texture information processing, researchers have explored the features critical for texture perception by examining metamers and linking them to hierarchical processing in the brain \citep{julesz1962visual, bergen1988early, freeman2013radial, freeman2013functional, okazawa2015image, ziemba2024neuronal}. For example, Freeman et al. \citep{freeman2013radial, freeman2013functional} identified a model capable of generating metamers at mid-level stages of visual processing, such as V2, using texture synthesis algorithms. The exploration of metamers has been applied not only to texture perception but also to other visual processes, such as peripheral vision \citep{rosenholtz2016capabilities, wallis2019image} and visual search \citep{rosenholtz2012summary}, as well as to other sensory modalities, including audition \citep{mcdermott2011sound} and haptics \citep{kuroki2021roles}.

The approach of using metamers for functional understanding has further advanced with the use of deep neural network models (DNNs) as artificial systems. Gatys et al. \citep{gatys2015texture} proposed an image generation technique that utilizes intermediate features in convolutional neural networks, summarizing them with statistical representations such as Gram matrices. This method has been particularly beneficial for cognitive neuroscience researchers interested in higher-order hierarchical visual processing. For example, in the context of object recognition research, a metamer-based approach combining this image algorithm with fMRI measurements has demonstrated that object-selective regions in higher ventral visual areas exhibit texture-like representation, computed by Gram matrices \citep{jagadeesh2022texture}. Furthermore, there have been attempts to evaluate the alignment between humans and models by assessing the extent to which metamers generated by DNN models align with human metamers \citep{feather2019metamers, feather2023model}.

\subsection{Challenges in Exploring High-Dimensional Human Metameric Spaces}

\begin{figure}[t]
    \centering
    \includegraphics[width=0.75\linewidth]{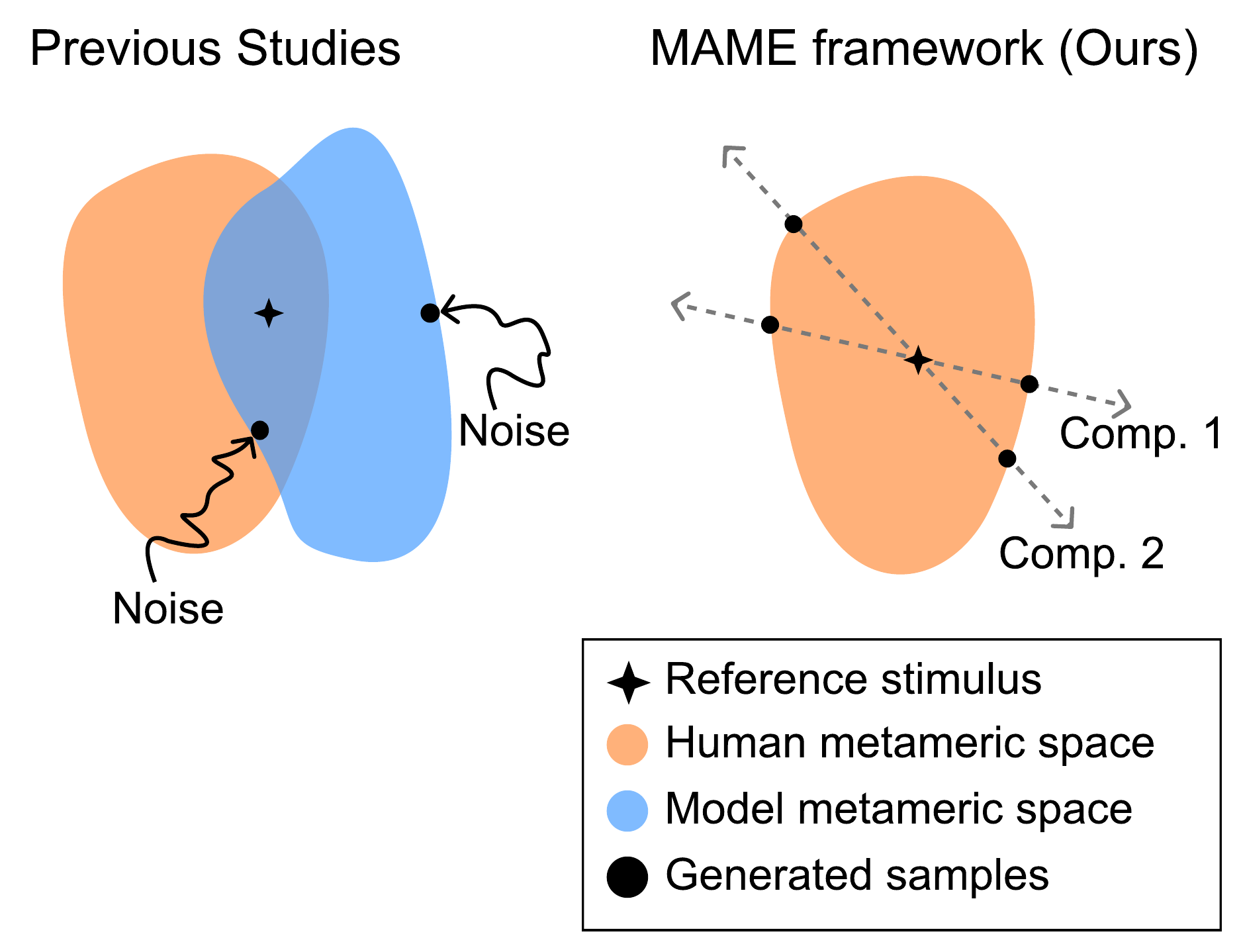}
    \caption{A conceptual illustration of the MAME framework. In previous studies, human metamers were explored indirectly by generating model metamers from noise images and presenting them to human observers. In contrast, our approach proposes a method for directly measuring human metamers by performing multidimensional exploration starting from a reference image.}
    \label{fig:concept}
\end{figure}

A key limitation of approaches that use deep neural networks (DNNs) to explore human metamers is that these approaches are inherently indirect. This limitation becomes evident when examining how previous studies have explored human metamers (Figure~\ref{fig:concept}).

In previous studies, model metamers are first synthesized from noise to match a target image in a given DNN, and human observers are then asked whether these images appear equivalent to the original target (Figure~\ref{fig:concept}, left). In this approach, the effectiveness of human metamer exploration critically depends on how well the model used for synthesis aligns with human visual perception. Indeed, adversarially trained convolutional neural networks (CNNs), which are known to exhibit human-like robustness across various tasks \citep{salman2020adversarially}, have been shown to generate metamers that better account for human peripheral vision \citep{harrington2021finding}.

However, this approach can fully reveal the structure of human metameric representations only if a suitably aligned model is available in advance. In this sense, human metamer exploration remains an indirect approach.
Instead, could we adopt a more direct approach to measuring human metamers? If such an approach were possible, we could first measure the human metameric space and then design models that align with it, ultimately achieving human-model functional alignment. In this case, rather than generating images from noise using a model, as in conventional studies, we could directly measure human metamers by freely exploring the human metameric space from a reference stimulus, as illustrated in Figure~\ref{fig:concept} (right). Furthermore, directly exploring human metamers is itself crucial for understanding human information processing, as exemplified by the discovery of color vision sensors.

Indeed, several previous studies have explored perceptual representations by probing distortions defined with respect to a reference stimulus. An example is the Maximum Differentiation (MAD) competition framework, which constructs stimulus pairs that maximize disagreement between competing computational models of perceptual quantities \citep{wang2008maximum, ma2016group, ma2018group}. Other approaches have further characterized the local sensitivity structure of models using analytically derived distortions, such as eigen- or principal-distortions, which approximate directions of maximal or minimal sensitivity in a representation space \citep{berardino2017eigen, feather2024discriminating}. In addition, recent studies have shown that deep generative models can synthesize stimuli that induce specific perceptual effects, such as visual illusions \citep{gomez2022synthesis}.

In many of these studies, the primary objective is to identify models that better approximate human perceptual judgments. To ensure experimental feasibility, these approaches typically reduce the stimulus space to a small set of dimensions. This strategy is effective for comparing candidate models and selecting those that align more closely with human perception. However, it is not primarily designed to reveal the intrinsic multidimensional structure of the human metameric space itself. Rather, the direct exploration of human metameric space requires a framework that enables multidimensional, online measurement of metameric boundaries guided by human perceptual feedback.

\subsection{Proposed Approach: The MAME Framework}

\begin{figure}[t]
    \centering
    \includegraphics[width=\linewidth]{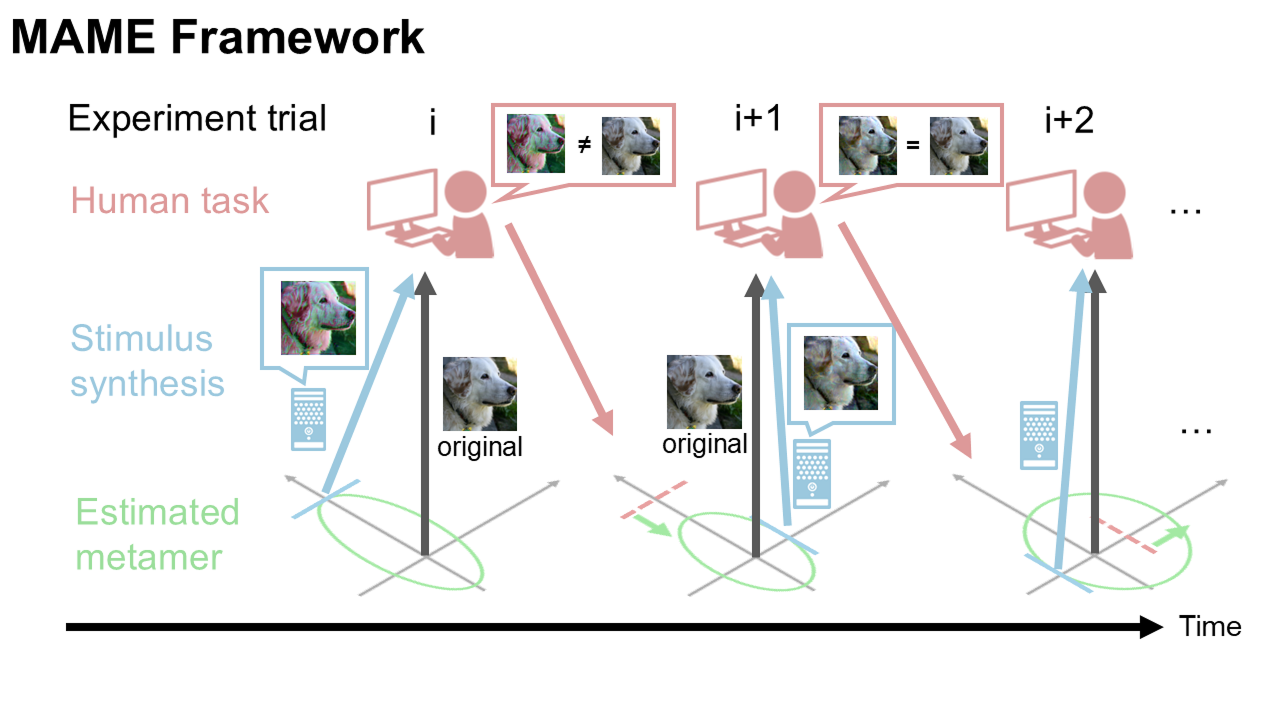}
    \caption{Exploring human metamers using the MAME framework.The exploration space is defined by the model (in this study, ICA components of CNN activations). Stimuli are synthesized along the defined directions, and their perceptual equivalence to a reference image is evaluated via a human behavioral task. Based on the feedback, the estimated human metamer is adaptively updated, and the task is repeated until convergence.}
    \label{fig:mame}
\end{figure}

In this study, we propose a framework for exploring high-dimensional human metameric spaces, called Multidimensional Adaptive Metamer Exploration (MAME) (Figure~\ref{fig:mame}). The MAME framework enables multidimensional exploration of human perceptual boundaries along model-defined directions by utilizing online image generation based on human feedback, thereby relaxing the constraints that a human-aligned model is needed beforehand. Specifically, in the MAME framework, the model is used solely to determine the search direction for human metamers. While it is better for the model to be biologically plausible when determining the direction, the final human metameric boundary can be determined independent of the model’s metamers. This aspect relaxes the constraints on human metamer exploration. Furthermore, to enhance the efficiency of multidimensional exploration, we employ online image generation guided by trial-wise human feedback, enabling the adaptive search for metameric boundaries.

In the experiment, we tested our MAME framework using a biologically plausible CNN, an adversarially trained ResNet50, on a natural image dataset, ImageNet. The key contributions of our MAME framework can be summarized as follows:

\begin{itemize}
    \item It can explore multiple model-defined directions relevant to human metameric perception simultaneously in one psychophysical experiment. 
    \item Using a guiding CNN, we found that the human discrimination sensitivity was lower for metameric images based on Gram-matrix representations derived from low-level CNN features than for those derived from high-level CNN features. This finding suggests a weaker alignment between low-level processing in humans and the model, highlighting a potential guideline for designing models that better align with human perception.
    \item The experimental environment is highly flexible because online image generation is performed on a Python server using a GPU, while the human experiment is conducted via a web browser with JavaScript.
\end{itemize}

The source code used in the experiments, as well as the analysis scripts required to reproduce the results, are available at Zenodo
(\url{https://doi.org/10.5281/zenodo.20789863}) and GitHub (\url{https://github.com/amano-k-lab/mame_hil}). In addition, the experimental results data are available at OSF (\url{https://doi.org/10.17605/OSF.IO/3TFHU}).

\section{Methods}
\label{methods}
\subsection{The MAME Framework}
The MAME framework is designed to adaptively explore the boundaries of the human metameric space. This exploration is achieved by systematically perturbing a given reference image in multiple dimensions (Fig. \ref{fig:concept}, right). The execution of the MAME framework relies on the following requirements:

\begin{enumerate}
    \item \textbf{Definition of the exploration direction of human metameric space}
    The human metameric space, initially defined in an input image space $\mathbb{R}^{W \times H \times C}$, must be compressed to a dimension practical for exploration (on the order of tens of dimensions) and defined using a coordinate system. Using CNNs or biologically plausible models coupled with compression methods, one can define the subspace direction to explore human metameric boundaries.

    \item \textbf{Online Stimulus Generation}  
    After determining the direction of exploration, an image-generation algorithm is required that is capable of generating stimuli at any position within the defined metameric space. The stimulus generation must be completed within a time frame that ensures compatibility with the human feedback process in the MAME framework. For instance, when incorporated into an ABX test \citep{freeman2011metamers}, this means generation within 2–3 seconds.

    \item \textbf{Feedback from Tasks performed by Humans or Animals}
    Finally, our MAME framework requires feedback from tasks performed by humans or animals. As described in the Introduction, metamers refer to images that are physically different but equivalent within a given system. However, the definition of “equivalence” changes across studies. Some adopt categorical equivalence based on behavioral responses \citep{feather2019metamers}, while others define it in terms of perceptual appearance \citep{freeman2011metamers,deza2017towards} using ABX tasks. Our method can be applied to feedback from any type of task if the computation speed of image generation is fast enough for real-time online experiments.
\end{enumerate}

Depending on the research objective, different methods can be selected for each requirement. In the present study, our aim was to demonstrate the effectiveness of MAME, a novel comprehensive framework, under specific conditions. To this end, we evaluated its effectiveness using a canonical approach within the context of metamer exploration research. Specifically, the MAME framework was implemented by (1) defining the direction of human metameric space using a CNN and ICA, (2) implementing a stimulus generation method that perturbs the reference stimulus via gradient descent, and (3) incorporating ABX tasks as feedback. The specific motivations for selecting each method are described in detail in the corresponding sections of the Methods. 

\subsection{Definition of the exploration direction of human metameric space}

In this section, we describe how the metameric search space was defined in the present study. We constructed the search space based on the activations of a robust variant of ResNet50 \citep{he2016deep} (ResNet50 robustness ImageNet L2-Norm $\epsilon = 3.0$) \citep{engstrom2019adversarial}. Among the many available DNN models, we selected this specific model because previous work has shown it to best approximate human metamers under indirect exploration paradigms \citep{feather2023model}. In our MAME framework, the model is used to define the direction of exploration. Therefore, in order to efficiently estimate human metamers under a limited number of conditions, it is preferable that the model be biologically plausible. In particular, we note that manipulating high-level features is more difficult in other models, as shown in the comparison results provided in the Supplementary Material (Fig. \ref{fig:model_compare}).

\begin{figure}[t]
    \centering
    \includegraphics[width=0.8\linewidth]{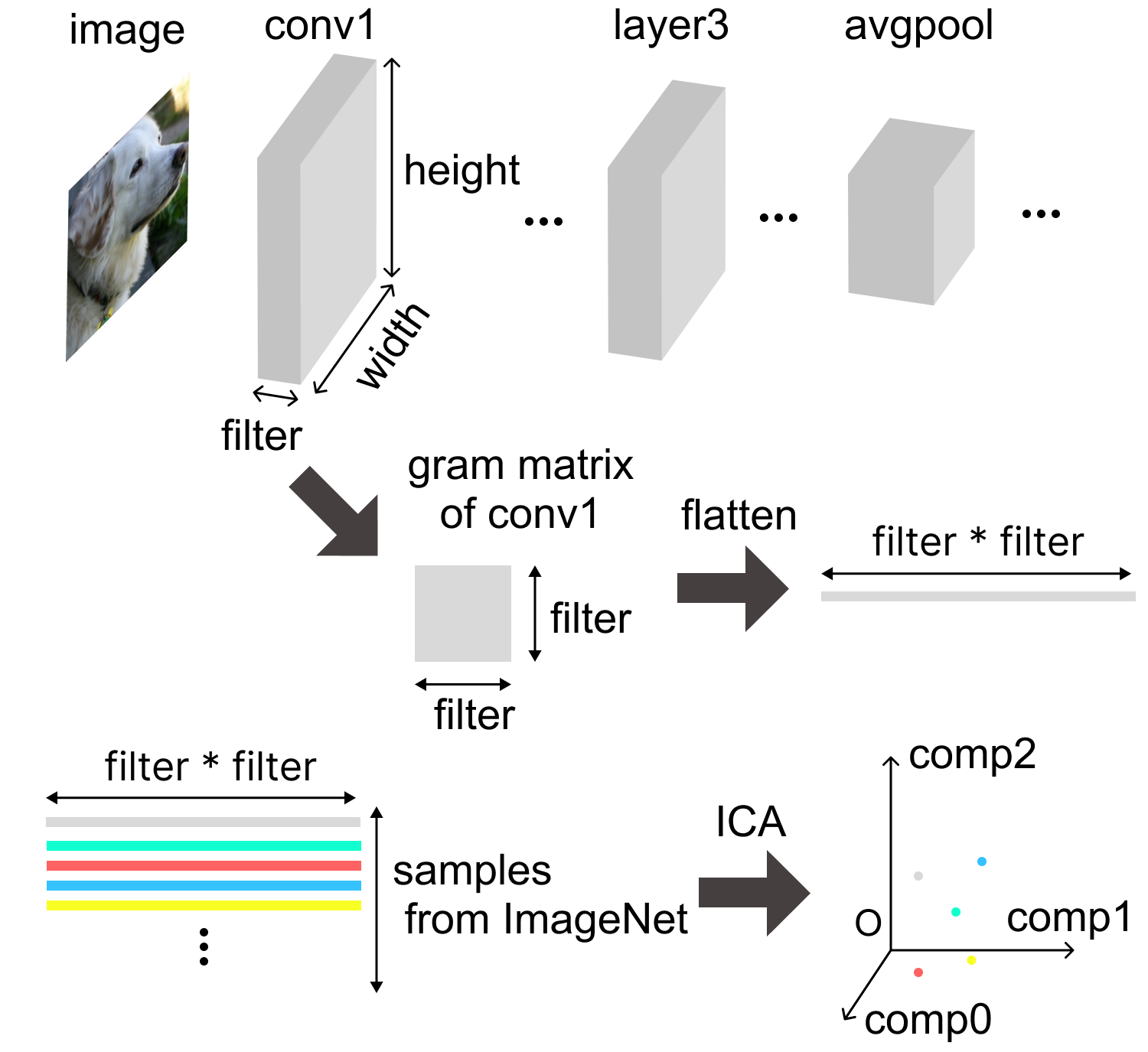}
    \caption{Definition of exploration directions in the present study.
An input image is passed through a robust ResNet50 model, and activations are extracted from the \texttt{conv1}, \texttt{layer3}, and \texttt{avgpool} layers. For each layer, filter-by-filter Gram matrices are computed and vectorized. Using these vectorized features from the ImageNet dataset, ICA is applied, and the top three components with the highest explained variance are selected as exploration axes for each layer.}
    \label{fig:metameric_space}
\end{figure}

Next, we detail how the search space was concretely defined (Fig.~\ref{fig:metameric_space}). First, natural images were passed through the robust ResNet50, while extracting Gram matrices from the activations of each layer as position-invariant features. Here, we used Gram matrices because the present study explored metamers based on the equivalence of perceptual appearance and focused summary statistics like textures representations \citep{gatys2015texture,gatys2016image,jagadeesh2022texture}.

A Gram matrix represents the similarity of activations across multiple filters, summed over spatial positions. Given the activation of the $l$-th layer of a CNN for an input image as $F_{jk}$, where $j$ denotes the filter index and $k$ denotes the spatial position, the Gram matrix $G_{ij}$ is defined as:

\begin{equation}
    \label{eq:gram}
    G_{ij} = \sum_k F_{ik} F_{jk}.
\end{equation}

In this study, Gram matrices were extracted from multiple layers of the model using 1,000 images randomly selected from the ImageNet validation dataset \citep{deng2009imagenet}.

After extracting the Gram matrix features, ICA was performed on them, and components with high explanatory power were selected, thereby considering images as points in the ICA coordinate space. The exploration direction was defined along the axes of ICA from the reference stimulus.

Although we used ICA to determine the exploration directions, several dimensionality reduction methods could be considered. In preliminary analyses, we also extracted components using PCA (Fig. \ref{fig:pca_ica}). However, the image patterns associated with each PCA component tended to be visually similar across components. For this reason, we adopted ICA for dimensionality reduction in the present analysis. In addition, ICA normalizes components to have unit variance, which facilitates comparisons of representations across different layers.

In the following, we describe in detail how dimensionality reduction of the metameric space was performed using ICA. ICA assumes non-Gaussianity of signals and computes a linear transformation to separate observed signals into statistically independent components. The Gram matrices of the target layers were vectorized to form feature vectors $\mathbf{x}$ for each image. Using the feature vectors of 1,000 images, an input data matrix $\mathbf{X} \in \mathbb{R}^{N \times D}$ was constructed, where $N=1000$ denotes the number of stimuli and $D$ the dimensionality of the flattened Gram features.

Prior to ICA, the data were centered and whitened. Whitening was performed using a PCA-based transformation, which decorrelates the data and normalizes the covariance structure. Specifically, if the covariance of the centered data is decomposed as $\Sigma = \mathbf{E}\mathbf{D}\mathbf{E}^\mathsf{T}$, the whitening transformation corresponds to
\[
\mathbf{X}_w = \mathbf{D}^{-1/2}\mathbf{E}^\mathsf{T}(\mathbf{X}-\boldsymbol{\mu}),
\]
which yields $\mathrm{Cov}(\mathbf{X}_w)=\mathbf{I}$.

ICA was then applied to the whitened data using FastICA (scikit-learn), yielding an unmixing matrix $\mathbf{W}$ such that
\begin{equation}
\label{eq:decomp}
\mathbf{S} = \mathbf{W}\mathbf{X}_w.
\end{equation}

With the unit-variance option, each recovered independent component has zero mean and unit variance across stimuli, i.e., $\mathrm{Cov}(\mathbf{S})=\mathbf{I}$. For a single stimulus $\mathbf{x}$, its ICA representation is obtained as $\mathbf{s}=\mathbf{W}\mathbf{x}_w$. The corresponding mixing matrix $\mathbf{A}$ allows reconstruction in the original feature space:
\[
\mathbf{X} \approx \mathbf{S}\mathbf{A}^\mathsf{T}.
\]

The number of independent components was pre-set to 100, and the top 3 components with the highest contribution were selected. The contribution of each component was quantified using the explained variance (EV) computed in the original (non-whitened) feature space as
\begin{equation}
\label{eq:ev}
\mathrm{EV}_i = 1 - \frac{\| \mathbf{X} - \hat{\mathbf{X}}_i \|_{\text{F}}^2}{\| \mathbf{X} \|_{\text{F}}^2},
\end{equation}
where
\begin{equation}
\label{eq:xhat}
\hat{\mathbf{X}}_i = \mathbf{S}_i \mathbf{A}^\mathsf{T},
\end{equation}
and $\mathbf{S}_i$ retains only the $i$-th independent component (Fig. \ref{fig:explained_variance}). Although ICA components have unit variance in the whitened space, their contributions differ in the original Gram feature space due to the mixing transformation.

Finally, we computed the metameric space based on features extracted from the \texttt{conv1}, \texttt{layer3}, and \texttt{avgpool} layers of the robust ResNet50. Each feature dimension is (64 x 64) for \texttt{conv1}, (1024 x 1024) for \texttt{layer3}, and (2048 x 2048) for \texttt{avgpool}. These layers were selected to examine early, intermediate, and higher-level stages of visual processing, while avoiding layers affected by skip connections in the ResNet architecture. Both positive and negative directions of each ICA component were used in the exploration.

\subsection{Generation of Target Images}
This section describes the method for generating images at designated positions using gradient descent. The image modification process was performed by specifying the target layer (\texttt{conv1}, \texttt{layer3}, and \texttt{avgpool}), the ICA component of interest (components 0, 1, 2), the direction of change (positive or negative), the magnitude of change (a positive scalar corresponding to the target value), and the reference image (i.e., the initial image of the modification).

The image generation procedure is as follows:

\begin{enumerate}
    \item Compute the ICA components of the reference image for the target layer as $\mathbf{y}_{\text{o}} = \mathbf{W}\mathbf{x}_{\text{o},w}$, 
    where $\mathbf{x}_{\text{o},w}$ is the whitened vectorized Gram matrix of the target layer. 
    \item Define the target component as \(\mathrm{p}\) and the target value as \( t \). The modified ICA component \( \mathbf{y}_{\text{t}}\) is given by:
    \begin{equation}\label{eq:ystep}
    \mathbf{y}_{\text{t}} = \mathbf{y}_{\text{o}} + t \mathbf{e}_\text{p},
    \end{equation}
    where $\mathbf{e}_\text{p}$ is a unit vector whose $p$-th element is 1 and 0 elsewhere. This operation modifies only the target component while keeping all others unchanged. If the change direction is negative, the unit vector is multiplied by $-1$.
    \item In each iteration, compute the ICA components of the current image as $\mathbf{y}_{\text{c}} = \mathbf{W}\mathbf{x}_{\text{c},w}$.
    \item Compute the mean squared error (MSE) between $\mathbf{y}_{\text{t}}$ and $\mathbf{y}_{\text{c}}$ as the loss function.
    \item Perform gradient descent using the Adam optimizer to minimize the loss and update pixel values of the current image while keeping the CNN model parameters fixed. 
    
\end{enumerate}

Image optimization was performed for 50 iterations using the Adam optimizer implemented in PyTorch~\citep{NEURIPS2019_bdbca288} (torch.optim.Adam, $\beta_1=0.9$, $\beta_2=0.999$, $\epsilon=10^{-8}$). 
The learning rate was determined as a function of the target value $t$. 
Specifically, for each layer, we first sampled a set of target values in log-space and optimized the learning rate independently for each value using Optuna~\citep{akiba2019optuna}. 
The resulting discrete set of optimal learning rates was then interpolated using cubic interpolation to obtain a continuous function of $t$.

To assess optimization fidelity, we computed the normalized final loss for each layer, component, and direction across 100 randomly sampled images. 
The normalized final loss remained consistently low across all conditions (Table~\ref{tab:final_loss}), indicating that the optimization converged reliably and that the main results are unlikely to be explained by differences in optimization fidelity.

\begin{figure}[t]
    \centering
    \includegraphics[width=\linewidth]{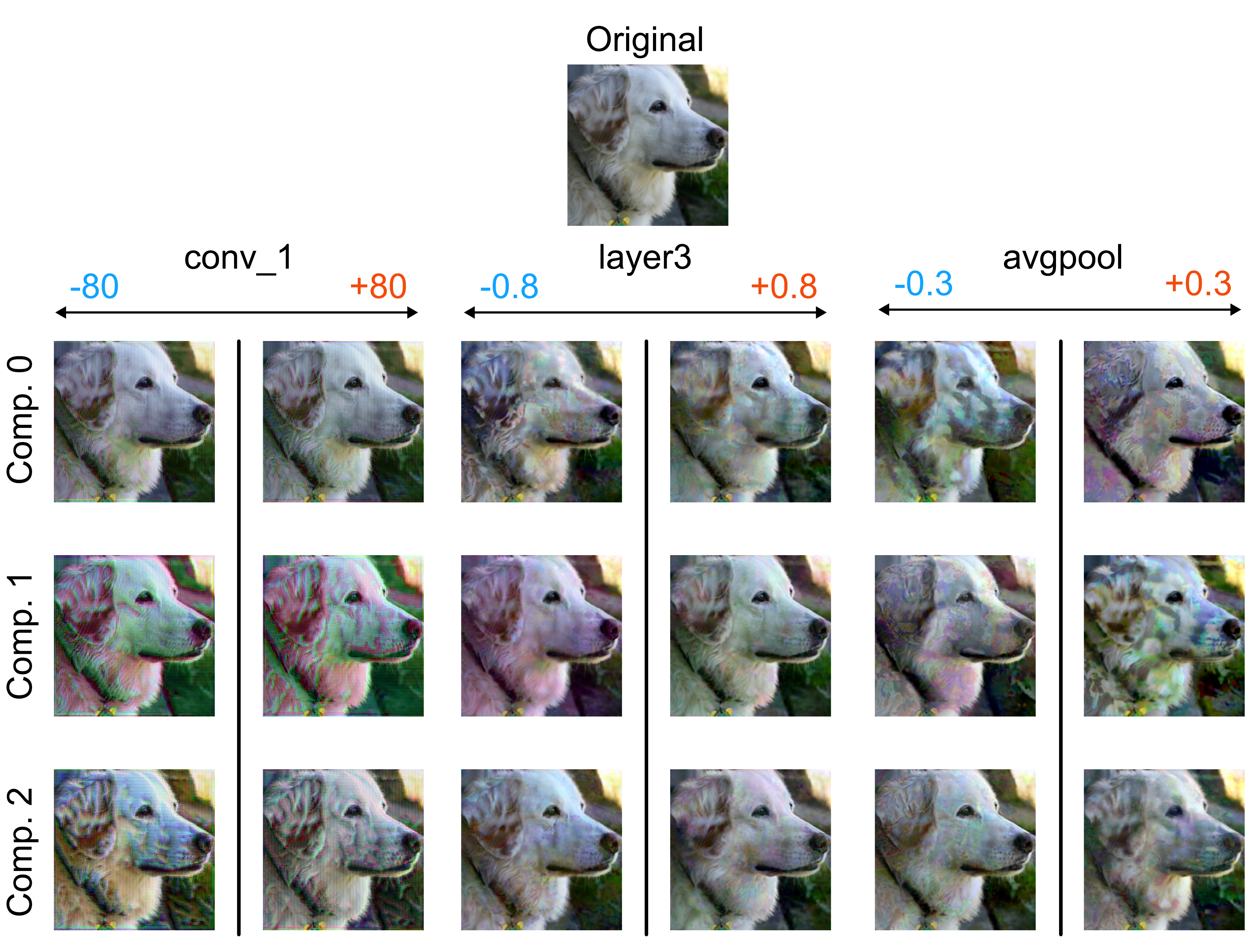}
    \caption{An example of the images used in the experiment, generated with three layers and three components each. The numbers (in red and blue) below the layer names represent the target ICA values used during image generation.The magnitude of modulation in this visualization is scaled according to the experimental results.}
    \label{fig:imggen_examples}
\end{figure}

Examples of image generation for the three layers and three components used in the experiment are shown in Fig.\ref{fig:imggen_examples} and Fig. \ref{fig:detail_example}. 

The selection of reference images for initialization was based on two criteria. First, we aimed to use images located near the zero point of the ICA components. Second, to prevent participants in the behavioral experiment from memorizing individual images, we needed to sample many images. For this purpose, from 50,000 images in the ImageNet validation dataset, we sampled images for three target layers whose ICA component distances from the origin were within the lowest 20\%. Among these, 614 images that met the selection criteria across all three layers were used as experimental stimuli (Table \ref{tab:ica_selection}).

\begin{table}[H]
\centering
\caption{ICA component criteria and the number of selected images}
\label{tab:ica_selection}
\begin{tabular}{|c|c|c|c|}
\hline
\textbf{Criterion*} & \textbf{Layer} & \textbf{Threshold} & \textbf{Number of Img} \\ \hline
\multirow{3}{*}{Lowest 20\%} & \texttt{conv1} & 1.22 & \multirow{3}{*}{614} \\ \cline{2-3}
                          & \texttt{layer3} & 1.73 &  \\ \cline{2-3}
                          & \texttt{avgpool} & 0.79 &  \\ \hline
\end{tabular}
\begin{flushleft}
\footnotesize{* An image satisfies a given criterion if its Euclidean distance in each layer is below the corresponding threshold.}
\end{flushleft}
\end{table}

\subsection{Feedback from Behavioral task}

As feedback in the present experiment, we employed an ABX task to measure perceptual similarity. While various forms of feedback could be considered—such as neural similarity or categorical judgments—our goal was to demonstrate the effectiveness of the novel MAME framework under specific conditions. In general, measuring perceptual appearance psychophysically is more time-consuming than categorical judgment in one trial. In contrast, computing the equivalence of neural or BOLD activations in neurophysiological or fMRI studies is a similar time course to experiments for perceptual appearance, although the experimental cost is much higher in neuroscience experiments. The ABX task was chosen because it offers a comparable trial duration to neural similarity experiments, while being less costly to implement. This makes it a suitable choice for assessing the feasibility of the MAME framework in an online experimental environment.

The experiment consisted of an ABX test comparing generated images with reference images. Based on the participants' response data, each image was updated until the exploration of human metamer boundaries converged.

Eight participants took part in the experiment. The study was conducted with approval by the Ethics and Safety Committee of the University of Tokyo. The MAME framework was implemented using a Python web framework (Django) and a browser-based experimental environment with JavaScript \citep{adolphe2022open}. Image generation was performed on a server machine with an NVIDIA RTX A2000 GPU, and the generated images were presented in the Google Chrome browser via a local server.

\begin{figure}[t]
    \centering
    \includegraphics[width=0.8\linewidth]{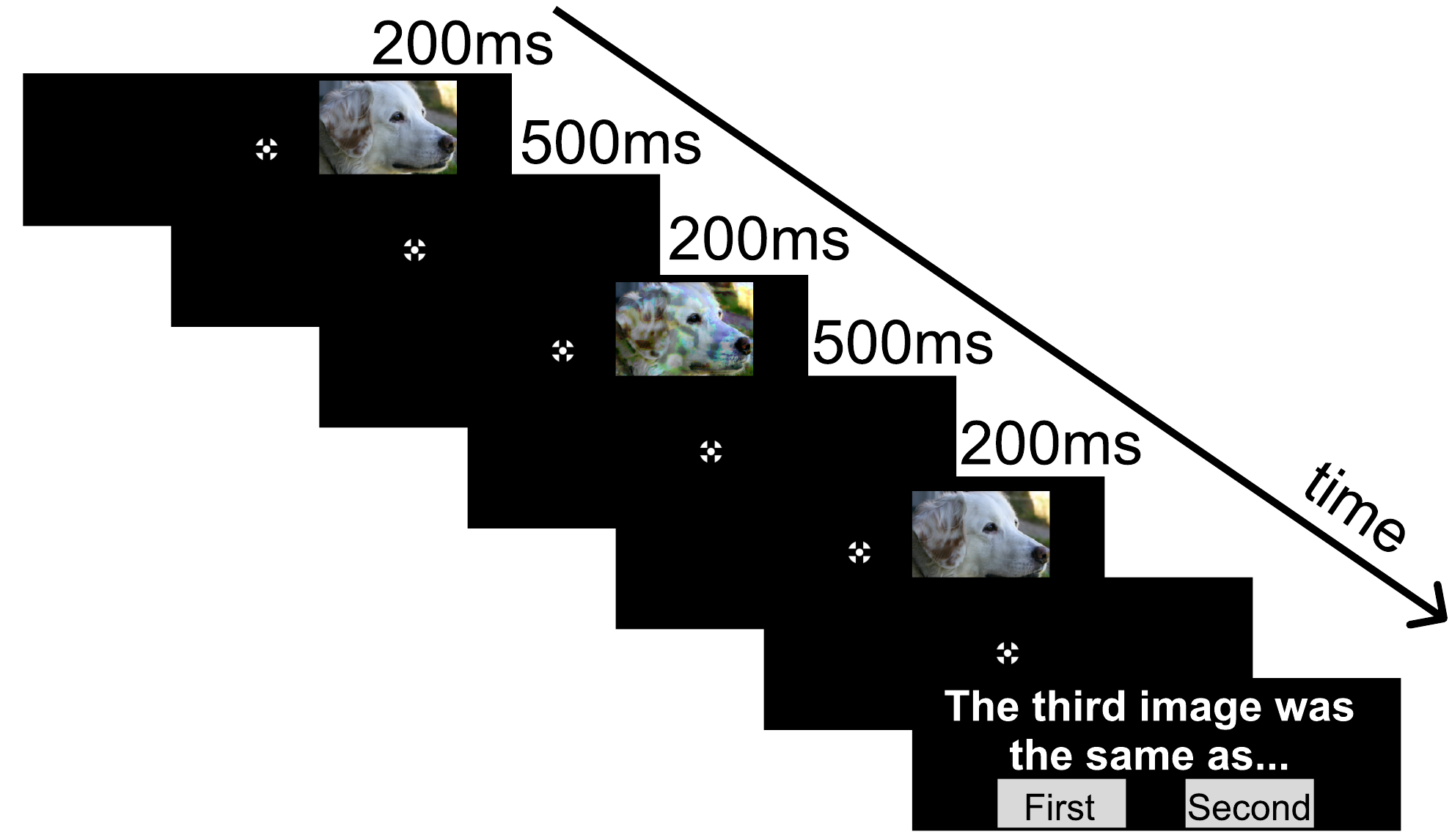}
    \caption{Overview of the ABX test. A 200 ms stimulus presentation was alternated with a 500 ms blank period.}
    \label{fig:ABXtest}
\end{figure}

The experiment was conducted in a dark room, with participants' heads fixed at a distance of 70 cm from the monitor. In all tasks, participants were instructed to fixate on a fixation point \citep{thaler2013best} at the center of the monitor (Fig.~\ref{fig:ABXtest}).

Furthermore, to ensure that participants were indeed fixating on the designated point during the experiment, gaze tracking was performed using the Tobii Pro Spark (Tobii, Sweden). Trials in which fixation was not maintained were not used for parameter updates.

The procedure of the ABX task is as follows (Fig.~\ref{fig:ABXtest}). In each trial, participants are presented with three stimuli in sequence: stimulus A, stimulus B, and stimulus X. The first two stimuli (A and B) are the reference and generated images in a random order. The third stimulus (X) is a test stimulus that is the same as A or B. Participants are required to identify whether X is the same as A or B. The presentation position was 4°, 8°, or 12° from the center depending on the eccentricity condition. The size of the presentation image was 4° x 4° in the visual angle.

Based on the results of each ABX test, the target value was updated using the staircase method (2-up 1-down rule). Specifically, if a participant correctly answered two consecutive trials in the ABX task, the boundary of the human metameric space was narrowed by one step. Conversely, if the participant made an error in a single trial, the boundary was shifted in the direction of expansion. Here, one step was defined as a fixed value for each layer, determined based on preliminary experiments: 10 for \texttt{conv1}, 0.3 for \texttt{layer3}, and 0.02 for \texttt{avgpool}. 

For the final threshold estimate, we fitted a sigmoidal psychometric function to the trial-by-trial responses collected during the staircase procedure using psignifit (2AFC task). The responses were treated as Bernoulli trials (correct/incorrect), and the threshold was defined as the stimulus level corresponding to 70.7\% correct, which matches the convergence point of the 2-up 1-down staircase.

The stimulus values generated during image synthesis did not always exactly match the predefined staircase step values. To quantify this discrepancy, we performed a post hoc analysis of all stimulus images used in the experiment. The RMSE between the target and actual stimulus values was 2.088 for conv1, 0.025 for layer3, and 0.005 for avgpool, all of which were substantially smaller than a single staircase step.

When fitting the psychometric functions, we therefore used the actual stimulus values rather than the predefined target staircase values. Specifically, for each staircase step, we computed the mean actual stimulus value across trials and fitted the psychometric function based on participants' responses to these stimulus levels. Examples of psychometric function fits are provided in the Appendix \ref{fig:psych_fit}.

In addition, because the stimulus values generated during image synthesis did not always exactly match the predefined staircase step values, we also estimated thresholds directly from the staircase trajectory by averaging the last five reversal points for each condition. This estimate reflects the actual stimulus levels presented during the experiment and was used as a complementary measure (see Appendix \ref{tab:overall_subjects}).

In summary, the experimental conditions consisted of three layers, three ICA components each explored in both positive and negative directions, and three levels of stimulus eccentricity (4°, 8°, and 12°). This resulted in a total of 54 conditions (3 × 3 × 2 × 3).

The experiment was conducted over two days for each participant. Each participant engaged in 15 experimental blocks, where each block consisted of 90 trials and took approximately 9–12 minutes to complete. In each block, the stimulus presentation position was fixed, and 18 conditions were presented five times each in a randomized order. Across the 15 blocks, three different stimulus presentation positions were used, with each position appearing in five blocks. The order of these positions was randomly determined at the start of the experiment. Following this protocol, each condition was tested 25 times using the ABX test. 

\section{Results}
\label{results}
The MAME framework was applied to each subject, and their metameric boundaries were obtained as ICA component values. 

Furthermore, analyses based on image metrics were conducted, allowing the metameric boundaries to be computed using these indices.

\subsection{Metameric Boundaries in ICA Components}
The metameric boundaries represented by ICA components, obtained directly from the experiment, are shown in Table~\ref{tab:pse_subjects} and Figure~\ref{fig:sup_plot_values}. As the effects of the ICA component and direction conditions were small (Fig. \ref{fig:sup_plot_values}), we first computed within-subject averages across these conditions. These mean values were averaged across different participants and shown in Table~\ref{tab:pse_subjects}.

\begin{table}[H]
    \centering
    \caption{Mean threshold values computed by averaging across ICA components, search directions, and subjects for each layer and eccentricity. The standard deviation represents inter-subject variability, computed from the threshold values that were first averaged within each participant over components and directions. ($n=8$).}
    
    \label{tab:pse_subjects}
    \begin{tabular}{lccc}
        \toprule
        Layer & Eccentricity (°) & Mean  & Std \\
        \midrule
        \texttt{conv1}    & 4  & 50.7380 & 9.9466 \\
                 & 8  & 67.4927 & 12.8030 \\
                 & 12 & 70.8025 & 9.6864 \\
        \midrule
        \texttt{layer3}   & 4  & 0.5124  & 0.1042 \\
                 & 8  & 0.6474  & 0.0631 \\
                 & 12 & 0.7165  & 0.0885 \\
        \midrule
        \texttt{avgpool}  & 4  & 0.1700  & 0.0219 \\
                 & 8  & 0.1701  & 0.0272 \\
                 & 12 & 0.1913  & 0.0269 \\
        \bottomrule
    \end{tabular}
\end{table}

To quantitatively evaluate the effects of layer and eccentricity on the metameric threshold, we performed a two-way repeated-measures ANOVA with \textit{Layer} (\texttt{conv1}, \texttt{layer3}, \texttt{avgpool}) and \textit{Eccentricity} ($4^\circ$, $8^\circ$, $12^\circ$) as within-subject factors. The analysis showed the significant main effects of \textit{Layer} and \textit{Eccentricity}
($F(2,14)=317.46$, $p<0.001$, $\eta_g^2=0.96$ and $F(2,14)=17.46$, $p<0.001$, $\eta_g^2=0.18$). For the main effect of \textit{Layer}, post-hoc pairwise comparisons with Bonferroni correction showed that the thresholds were significantly higher in the order of \texttt{conv1}, \texttt{layer3}, and \texttt{avgpool} ($p<0.001$). The interaction was also significant
($F(4,28)=16.71$, $p<0.001$, $\eta_g^2=0.30$). Based on the significant interaction, we further examined the simple effects of eccentricity within each layer.

For the \texttt{conv1} layer, thresholds increased significantly with eccentricity ($F(2,14)=16.96$, $p<0.001$, $\eta_g^2=0.39$). Post-hoc pairwise comparisons with Bonferroni correction showed that thresholds at $8^\circ$ and $12^\circ$ were significantly higher than those at $4^\circ$, whereas the difference between $8^\circ$ and $12^\circ$ was not significant. For the \texttt{layer3} layer, the effect of eccentricity was also significant ($F(2,14)=14.46$, $p<0.001$, $\eta_g^2=0.49$). Pairwise comparisons revealed that thresholds at $8^\circ$ and $12^\circ$ were significantly higher than those at $4^\circ$, while the difference between $8^\circ$ and $12^\circ$ was not significant. For the \texttt{avgpool} layer, the effect of eccentricity was weaker but still significant ($F(2,14)=5.37$, $p=0.016$, $\eta_g^2=0.13$). However, pairwise comparisons showed that none of the pairs were statistically significant ($p>0.05$).

The observed differences in thresholds across layers suggest that the sensitivity of the metameric boundary defined by \texttt{conv1} is lower than that of \texttt{layer3} and \texttt{avgpool}. However, this result should be interpreted carefully when considering ICA dimensionality selection. When ICA is applied to the features of each layer, whitening is performed with respect to the natural image distribution, implying that the values of the ICA components are scaled within that distribution for each layer. Nevertheless, as shown in Fig.~\ref{fig:explained_variance}, the proportion of variance in the original Gram matrix explained by the independent components differs substantially across layers. This suggests that comparing sensitivity across layers requires evaluating the threshold images using multiple image metrics in addition to ICA-based thresholds. We address this in the following section.

\subsection{Evaluation by multiple image metrics}

We examined how much the threshold images obtained under each experimental condition differed from the original images using several image quality metrics. Here, the threshold images obtained for each condition can be interpreted as being equally different from the original image for human observers across different stimulus dimensions. If the directions explored by a model are well aligned with human metamers, smaller perturbations from the original image are expected to be sufficient for detection by human observers. This is because even a slight perturbation breaks the metameric equivalence for the model, meaning that the perturbed image is no longer a metamer from the model’s perspective.

\begin{figure}[t]
    \centering
    \includegraphics[width=\linewidth]{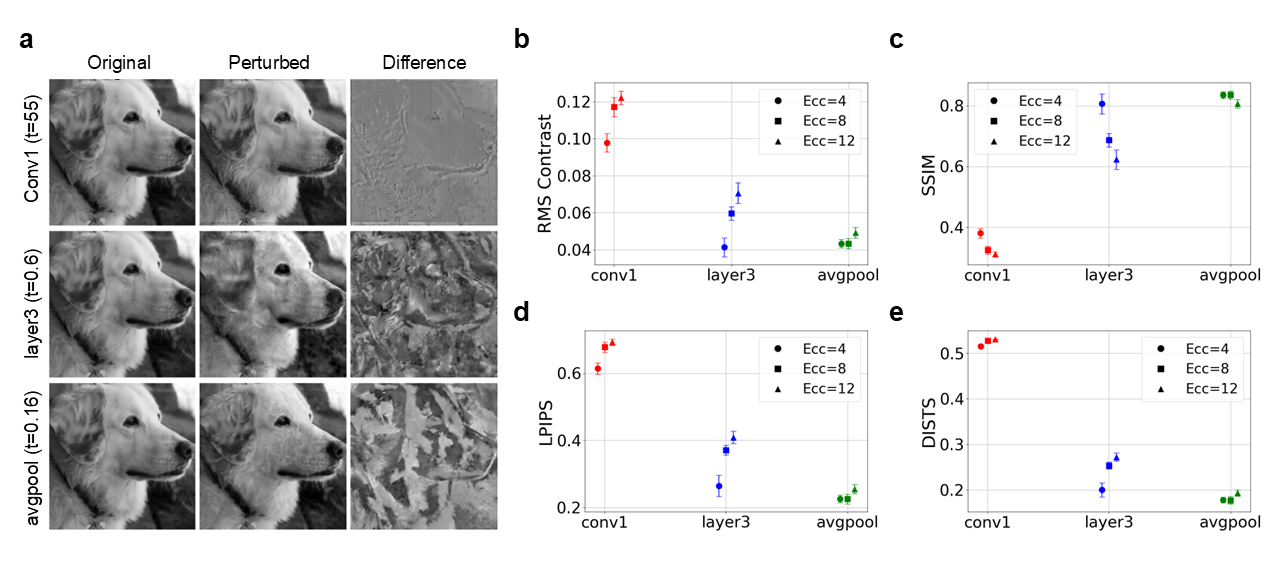}
    \caption{(a) Examples of threshold images for each layer (displayed in grayscale). The target value \( t \) is determined based on the threshold for each layer. The rightmost column shows the difference images. (b) The RMS contrast of the difference image was computed. A total of $n=30$ images were generated based on the threshold values obtained from each participant in the behavioral experiment and were averaged across images. The RMS contrast values were then averaged across participants, and the mean and standard error are plotted. Lower values indicate smaller contrast differences. Similarly, (c) SSIM, (d) LPIPS, and (e) DISTS between the original and perturbed images were computed, and the mean and standard deviation are plotted. Higher values for SSIM and lower values for LPIPS and DISTS indicate higher similarity between the original and perturbed images.}
    \label{fig:RMS_linear}
\end{figure}

First, we conducted an evaluation using RMS contrast. RMS contrast is an index that quantifies the overall contrast strength of a grayscale image. In the analysis, for each layer and eccentricity, grayscale difference images were generated by subtracting the reference image from the perturbed image using the target ICA component values at the experimentally determined thresholds (Figure~\ref{fig:RMS_linear}a). The RMS contrast of these difference images was then computed. Each threshold image was computed separately for each participant.

The results of the RMS analysis are shown in Figure~\ref{fig:RMS_linear}b. As in the previous section, we performed a two-way repeated-measures ANOVA with \textit{Layer} (\texttt{conv1}, \texttt{layer3}, \texttt{avgpool}) and \textit{Eccentricity} ($4^\circ$, $8^\circ$, $12^\circ$) as within-subject factors. The analysis revealed significant main effects of \textit{Layer} and \textit{Eccentricity}
($F(2,14)=428.12$, $p<0.001$, $\eta_g^2=0.87$; $F(2,14)=40.70$, $p<0.001$, $\eta_g^2=0.36$). The interaction was also significant
($F(4,28)=4.84$, $p<0.001$, $\eta_g^2=0.15$).

Simple effects analyses with Bonferroni-corrected pairwise comparisons showed that, for both \texttt{conv1} and \texttt{layer3}, the values at $4^\circ$ were significantly lower than those at $8^\circ$ and $12^\circ$ ($p<0.05$, Bonferroni-corrected), whereas no significant differences were found between $8^\circ$ and $12^\circ$ ($p>0.05$, Bonferroni-corrected). In contrast, for the \texttt{avgpool} layer, no significant differences were found across eccentricities ($p>0.05$, Bonferroni-corrected).

For comparisons across layers, the same analyses revealed that only the difference between \texttt{layer3} and \texttt{avgpool} at $4^\circ$ was not statistically significant ($p>0.05$, Bonferroni-corrected), whereas all other between-layer comparisons showed significant differences ($p<0.05$, Bonferroni-corrected). Consistent with the ICA threshold analysis, these results suggest that human sensitivity is lower when the metameric boundary is explored using lower-level image features, such as those in \texttt{conv1}.

Next, in addition to RMS contrast, we evaluated the threshold images using several other image quality metrics. Specifically, we employed the Structural Similarity Index (SSIM)\citep{wang2004ssim}, LPIPS \citep{zhang2018unreasonable}, and DISTS \citep{ding2020image}. SSIM assesses local structural similarity within a sliding window. LPIPS and DISTS are recent perceptual metrics that capture high-level feature similarities based on deep neural network representations. For LPIPS and DISTS, evaluations were performed using the original color images.

The results of these analyses are shown in Figure~\ref{fig:RMS_linear}c--e. Across all metrics, the overall trends were consistent with those observed for RMS contrast and the ICA-based thresholds. Note that for SSIM, higher values indicate greater similarity to the original image. The results of the two-way repeated-measures ANOVA for each metric are summarized in Tables~\ref{tab:anova_results1}, ~\ref{tab:anova_results2}, and ~\ref{tab:anova_results3}. The conditions showing significant effects in the ANOVA were consistent with those observed for RMS contrast.

Taken together with the ICA threshold results, these findings suggest that the Gram-matrix representations derived from \texttt{conv1} are less well aligned with human sensitivity than those derived from \texttt{layer3} and \texttt{avgpool}. Furthermore, threshold images generated from higher-level features, such as \texttt{avgpool}, appear to be less affected by eccentricity.

\section{Discussion}
\label{discussion}
\subsection{Summary of Main Findings}
The present study proposed the MAME framework, an experimental framework to directly explore human metamers using online feedback by humans with weak supervision of artificial models. The effectiveness of this framework was demonstrated by successfully estimating multi-dimensional image metamer boundaries. Experimental results indicated that, among image generation layers, the threshold in \texttt{conv1} was higher based on the magnitude of ICA thresholds, RMS contrast, SSIM, LPIPS, and DISTS, suggesting that changes in this layer were less noticeable compared to others. Furthermore, the effect of eccentricity on sensitivity was less pronounced for higher-level features.

To contextualize our findings, it is important to revisit what model metamers represent in the current study. In our MAME framework, when an image is shifted even slightly in the model’s exploration direction from the reference image, the shifted image is no longer a model metamer for the model. This is because exploration is carried out in the model’s activation space, where model metamers are defined as stimuli that are distinct in image space but equivalent in the target activation space. As a result, a single step in this space exceeds the model’s metameric boundary in image space.

This relationship implies that lower mean thresholds observed in human responses at a given hierarchical level indicate a closer alignment between the metamer boundaries of the human and the model. Given this relationship, the experimental results showed that threshold images generated from \texttt{conv1} exhibited higher ICA thresholds than those from \texttt{layer3} and \texttt{avgpool}. Moreover, evaluations based on image quality metrics indicated that these differences cannot be readily explained by differences in the scale of ICA components across layers. These findings suggest that Gram-matrix representations derived from lower-level CNN layers are less well aligned with human perception.

\subsection{Alignments of Higher and Lower Visual Processing in Humans and Models}

The finding related to the alignment of higher-level image features is consistent with recent studies using robust CNNs \citep{feather2019metamers,feather2023model,gaziv2024strong}. For instance, it has been shown that adversarial attacks on high-level activations can induce category modulations that are also perceptible to humans \citep{gaziv2024strong}. Furthermore, the alignment of high-level features has been a topic of intense discussion in both visual and LLM literature. For example, the Platonic Representation Hypothesis \citep{huh2024platonic} proposes that 'neural networks, trained with different objectives on different data and modalities, are converging to a shared statistical model of reality in their representation spaces.' From this perspective, the strong alignment of high-level representations between humans and modern neural models, trained with various data, might be a straightforward consequence.

However, our experimental results also suggest a functional misalignment between models and humans in low-level processing. Given that convolutional neural networks (CNNs) are originally inspired by the information processing of simple and complex cells in the primary visual cortex \citep{fukushima1980neocognitron}, this result may seem counterintuitive. One reason is that the first-layer filters in ResNet50 have a high-resolution size of 7×7 pixels and lack low-frequency components.

Indeed, previous work by Berardino and colleagues demonstrated that when perturbations aligned with directions of maximal or minimal model sensitivity (eigen-distortions) were applied to reference stimuli, models that mimicked early visual processing stages, such as the lateral geniculate nucleus (LGN), showed closer alignments to human perceptual sensitivity than standard CNN architectures \citep{berardino2017eigen}. Furthermore, attempts to incorporate such early-stage processing into CNNs have been reported to improve model robustness and generalization \citep{dapello2020simulating}. Taken together, these findings and our present results suggest that reconsidering the early-stage architecture of models may be crucial for improving functional alignment between humans and artificial systems.

\subsection{Model-Guided Distortions, Limitations, and Future Applications}

Several previous studies have investigated human perceptual sensitivity by modulating images with respect to a reference stimulus. For example, eigen-distortions and principal distortions \citep{berardino2017eigen, feather2024discriminating} use the Fisher information matrix to characterize the local geometry of a model’s representation and to quantify how sensitivity to specific perturbation directions compares between models and human observers. These approaches are typically designed to identify models whose local sensitivity structure better approximates human perception. To ensure experimental feasibility, they often evaluate sensitivity along a relatively small number of analytically defined directions.

A key difference between these previous approaches and our framework is how the exploration directions are defined. Rather than deriving perturbation directions from local sensitivity structures around a particular image, we derive exploration directions from independent components estimated across a large set of natural images. This approach enables the extraction of multiple exploration directions from a single reference image and facilitates multidimensional exploration within a practical experimental setting. Given that human metameric representations are likely to be multidimensional, we believe that this strategy provides a reasonable starting point for exploring candidate metameric boundaries.

Although Gram-matrix-based ICA is not the only possible strategy, it offers several practical advantages. By performing ICA over natural image distributions, the resulting directions reflect dominant statistical variations across images and can partially mitigate differences in scale across hierarchical layers. This property is particularly beneficial when comparing modulation effects across layers with inherently different activation magnitudes.

At the same time, the effectiveness of this approach may differ across levels of visual representation. If perceptual sensitivity at early stages of visual processing is more image-dependent than at later stages, exploration directions estimated across many images may be less effective for characterizing sensitivity in early representations, whereas directions derived from local sensitivity structures around individual images may provide a more appropriate description. Future work should investigate how multidimensional exploration approaches such as MAME can be integrated with image-specific local sensitivity measures, including those derived from Fisher information matrices.

More generally, the explored stimulus conditions in the present study do not comprehensively cover all possible directions of human metamer space. Furthermore, because the ICA components were estimated from only 1,000 images in a high-dimensional feature space, larger and more diverse image sets may reveal additional informative directions. Exploring alternative feature representations, image datasets, and computational models will be important for obtaining a more complete characterization of human metameric representations.

Despite these limitations, our approach provides a flexible framework for studying perceptual representations. Because the framework can be executed in a web browser, it is largely free from hardware constraints and can be easily combined with a wide range of experimental environments, including neuroimaging settings such as fMRI as well as large-scale crowdsourcing experiments. Therefore, we believe that MAME can serve as a useful tool for future investigations of perceptual representations across diverse domains and methodologies.

\section{Conclusions}
\label{conclusions}
In this study, we proposed the Multidimensional Adaptive Metamer Exploration (MAME) framework, which combines hierarchical neural network–based image generation with human-in-the-loop exploration, to directly map the multidimensional human metameric space. Our evaluation experiment demonstrated that MAME can measure this space within a single psychophysical experiment. The results further revealed that a recent human-aligned model exhibits a substantial functional mismatch between human perception and Gram-matrix representations derived from lower-level CNN layers, despite closer alignment for representations derived from higher layers. Because the user interface of MAME operates entirely within a web browser, it can be applied in fMRI environments and large-scale crowdsourcing experiments, making it a diverse tool for exploring metamers in human vision and for advancing interpretable AI.

\section{Acknowledgments}
This work was supported by JSPS KAKENHI Gran Numbers JP21H04909 and JP26K02927. We used ChatGPT (OpenAI) for English language editing. All content was reviewed and verified by the authors.

\bibliographystyle{apacite} 
\bibliography{references}  

\clearpage

\section{Supplementary Materials}
\label{supplementary materials}
\renewcommand{\thefigure}{S\arabic{figure}}
\setcounter{figure}{0} 
\renewcommand{\thetable}{S\arabic{table}}
\setcounter{table}{0} 
\begin{figure}[H]
    \centering
    \includegraphics[width=0.8\linewidth]{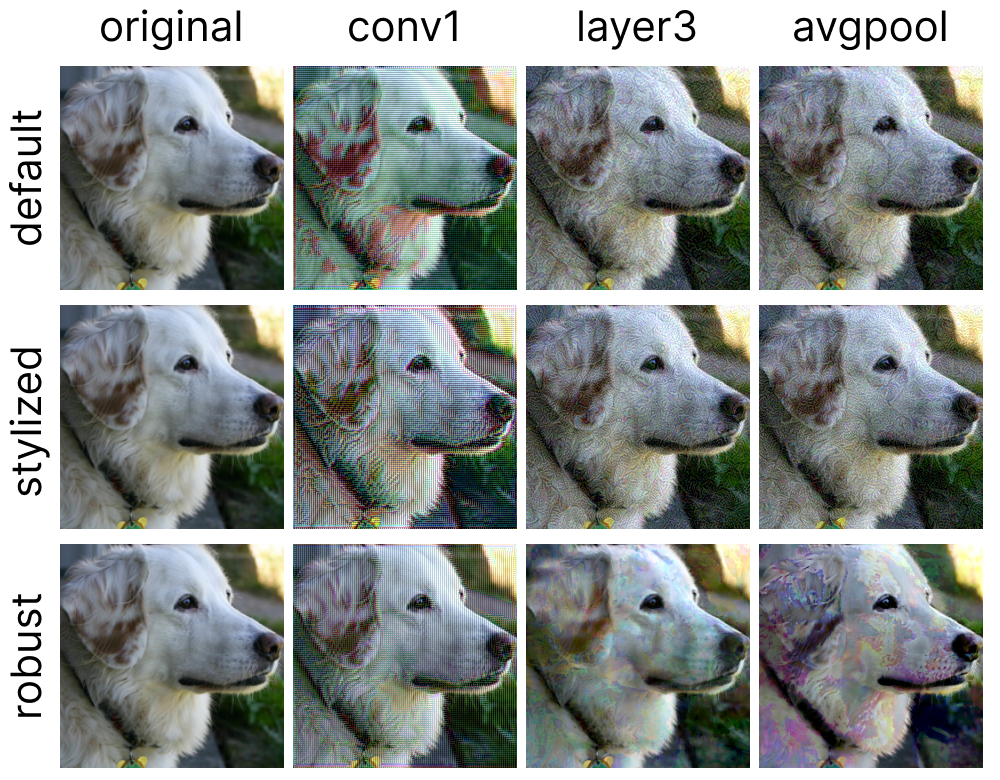}
    \caption{Examples of image modulation using models with different training regimes but the same architecture as the robust ResNet50 used in this study.
    The robust ResNet was trained to be resistant to adversarial attacks, whereas the default model was trained on standard ImageNet data, and the stylized model was trained on stylized versions of ImageNet \citep{geirhos2018imagenet}. When the same modulation parameters were applied to each model, we found that, particularly in higher layers, the default and stylized models failed to produce clear modulations from the original image, in contrast to the robust model (cf. Feather, 2023).}
    \label{fig:model_compare}
\end{figure}
\newpage

\begin{figure}[H]
    \centering
    \includegraphics[width=0.7\linewidth]{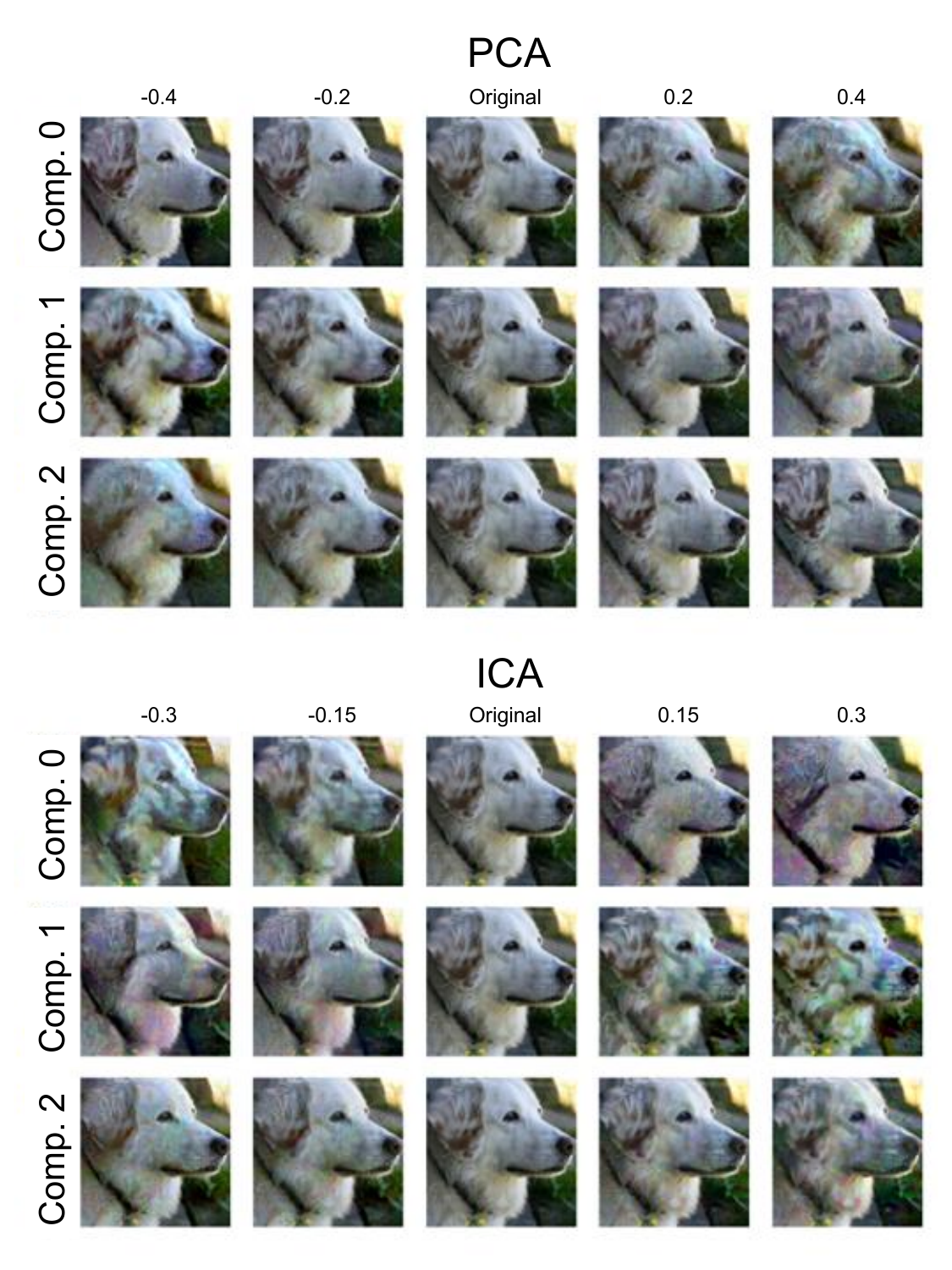}
    \caption{Comparison of image modulation using PCA and ICA in high-level representations (computed at the \texttt{avgpool} layer of the robust ResNet50). (Upper) Three principal components were extracted via PCA from vectorized Gram matrices, and image modulation was performed along each component axis. (Bottom) Image modulation using ICA, as employed in the present study. Due to the learning properties of DNNs, high-level features are expected to contain many independent components. The PCA-based modulation, which preserves orthogonality in the Gram matrix space, fails to capture such diversity, making it difficult to generate varied image modulations.}
    \label{fig:pca_ica}
\end{figure}
\newpage

\begin{figure}[H]
    \centering
    \includegraphics[width=1\linewidth]{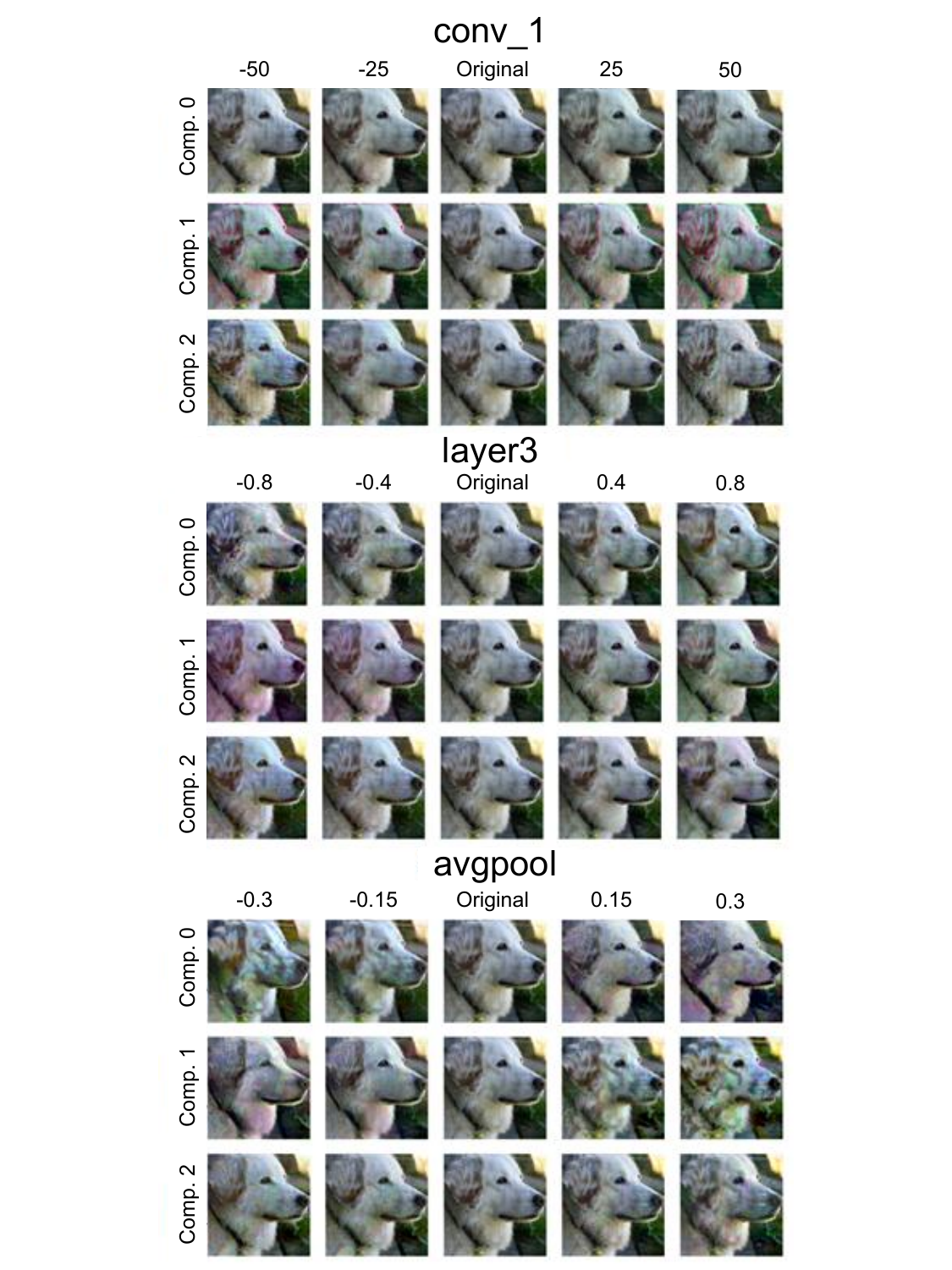}
    \caption{ Examples of image modulation for each layer and each component. This figure presents a finer-step version of Fig. \ref{fig:imggen_examples}.}
    \label{fig:detail_example}
\end{figure}
\newpage

\begin{figure}[H]
    \centering
    \includegraphics[width=0.5\linewidth]{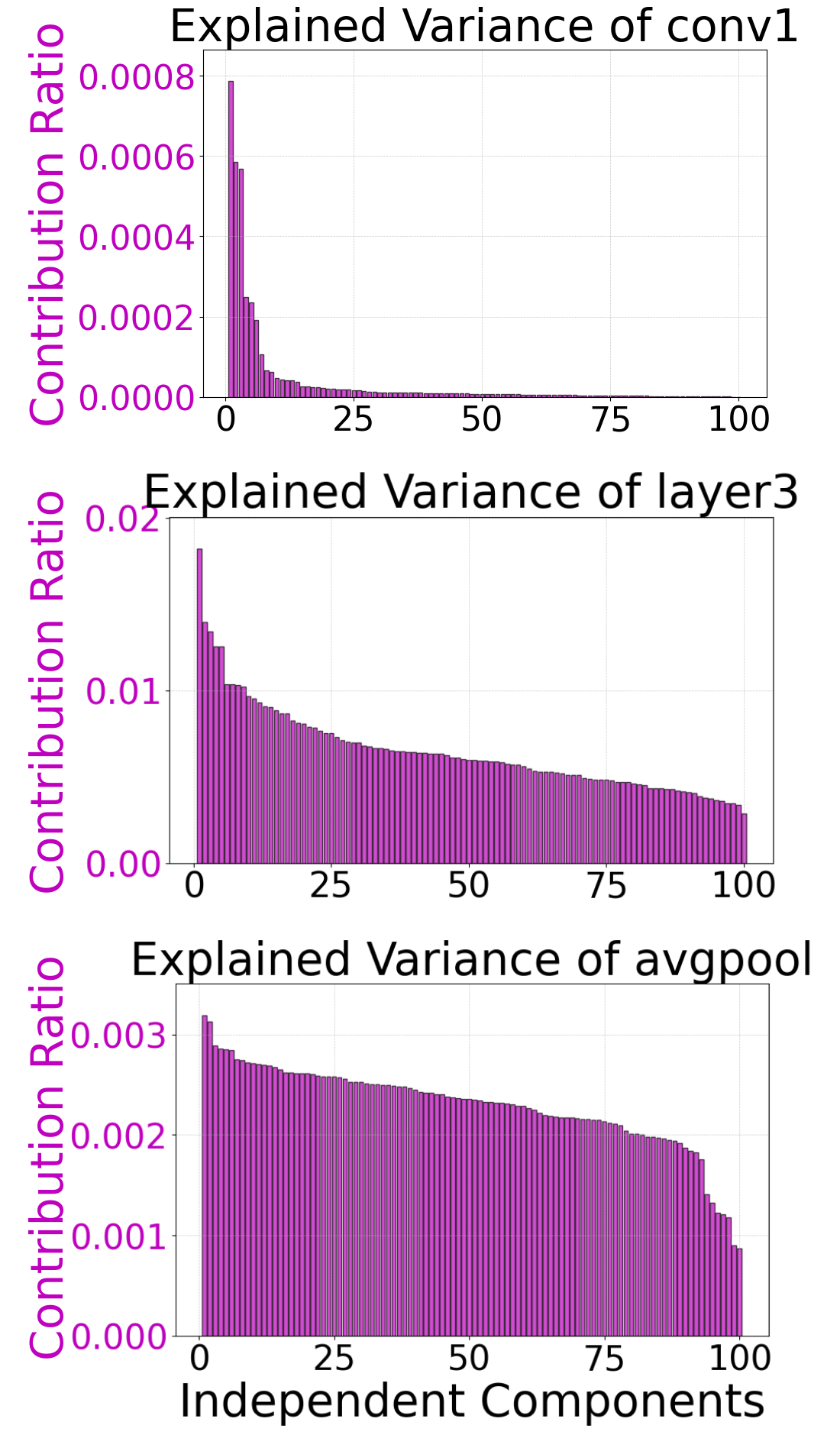}
    \caption{Explained variance of ICA components sorted in descending order. (Top) \texttt{conv1}, (Middle) \texttt{layer3}, (Bottom) \texttt{avgpool}.}
    \label{fig:explained_variance}
\end{figure}
\newpage

\begin{table}[t]
\centering
\caption{
Normalized final loss for each layer, ICA component, and direction (mean $\pm$ SD, $n=100$).
The normalized final loss is defined as the ratio between the final loss and the initial loss before optimization.
For each layer, target values were sampled uniformly in log-space (i.e., log-spaced sampling) within the specified $\log_{10}(t)$ range.
}
\label{tab:final_loss}

\begin{tabular}{lcccc}
\toprule
Layer & $\log_{10}(t)$ range & Component & Direction & Normalized final loss \\
\midrule
\texttt{conv1} 
    & [0.0, 3.0]
    & 0 & + & 0.006636 $\pm$ 0.024610 \\
    &       &   & $-$ & 0.007039 $\pm$ 0.027097 \\
    &       & 1 & + & 0.004328 $\pm$ 0.010553 \\
    &       &   & $-$ & 0.004179 $\pm$ 0.010673 \\
    &       & 2 & + & 0.005512 $\pm$ 0.010508 \\
    &       &   & $-$ & 0.007136 $\pm$ 0.014643 \\
\midrule
\texttt{layer3} 
    & [-2.0, 1.0]
    & 0 & + & 0.006004 $\pm$ 0.012682 \\
    &       &   & $-$ & 0.005656 $\pm$ 0.010533 \\
    &       & 1 & + & 0.007383 $\pm$ 0.014922 \\
    &       &   & $-$ & 0.006431 $\pm$ 0.014312 \\
    &       & 2 & + & 0.005035 $\pm$ 0.008581 \\
    &       &   & $-$ & 0.006483 $\pm$ 0.011846 \\
\midrule
\texttt{avgpool} 
    & [-2.0, 1.0]
    & 0 & + & 0.013961 $\pm$ 0.055785 \\
    &       &   & $-$ & 0.011262 $\pm$ 0.044577 \\
    &       & 1 & + & 0.006213 $\pm$ 0.021678 \\
    &       &   & $-$ & 0.007241 $\pm$ 0.016612 \\
    &       & 2 & + & 0.008365 $\pm$ 0.026241 \\
    &       &   & $-$ & 0.006456 $\pm$ 0.017010 \\
\bottomrule
\end{tabular}
\end{table}

\begin{table}[H]
    \centering
    \caption{Threshold estimates based on staircase reversal points. In addition to the psychometric-function analysis reported in Table~\ref{tab:pse_subjects}, thresholds were estimated by averaging the last five reversal points of the staircase trajectory for each condition.}
    \label{tab:overall_subjects}

    \begin{tabular}{lccc}
        \toprule
        Layer & Eccentricity (°) & Mean  & Std \\
        \midrule
        \texttt{conv1}    & 4  & 47.915 & 8.8166 \\
                 & 8  & 63.0404 & 11.8613 \\
                 & 12 & 63.7918 & 8.5024 \\
        \midrule
        \texttt{layer3}   & 4  & 0.486  & 0.1275 \\
                 & 8  & 0.5983  & 0.0686 \\
                 & 12 & 0.6407  & 0.0533 \\
        \midrule
        \texttt{avgpool}  & 4  & 0.1538  & 0.0182 \\
                 & 8  & 0.153  & 0.0253 \\
                 & 12 & 0.1688  & 0.0203 \\
        \bottomrule
    \end{tabular}
\end{table}

\newpage

\begin{figure}[H]
    \centering
    \includegraphics[width=\linewidth]{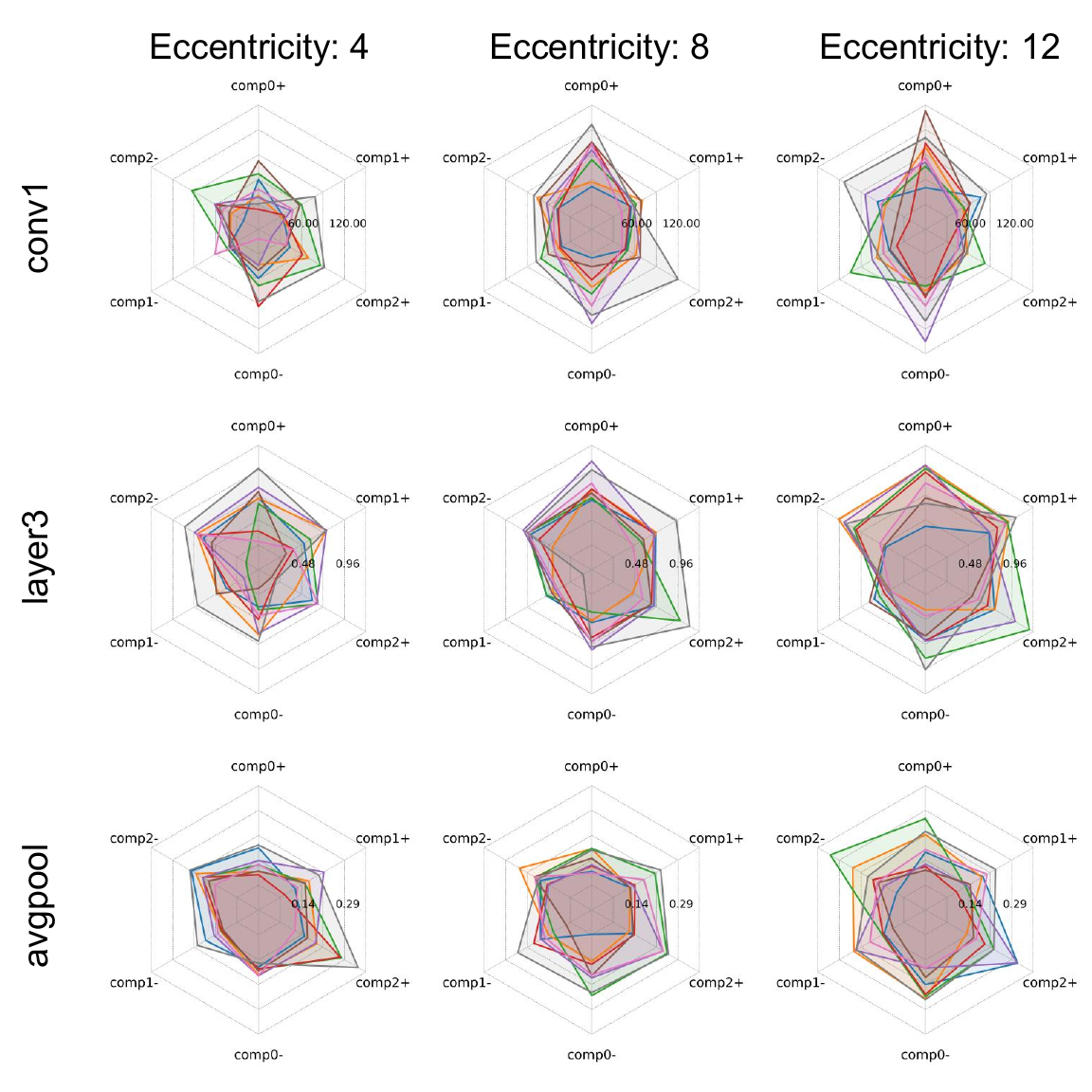}
    \caption{The ICA threshold values obtained from all subjects ($n=8$) across nine radar charts. Each row corresponds to a convolutional layer, and each column represents the eccentricity (in degrees) at which the stimulus was presented. The ICA threshold value at each point indicates the mean of reversal points obtained through the staircase method ($n \geq 6$). Colors distinguish individual subjects, with the same color consistently representing the same subject across all charts.}
    \label{fig:sup_plot_values}
\end{figure}

\newpage
\begin{figure}[H]
    \centering
    \includegraphics[width=\linewidth]{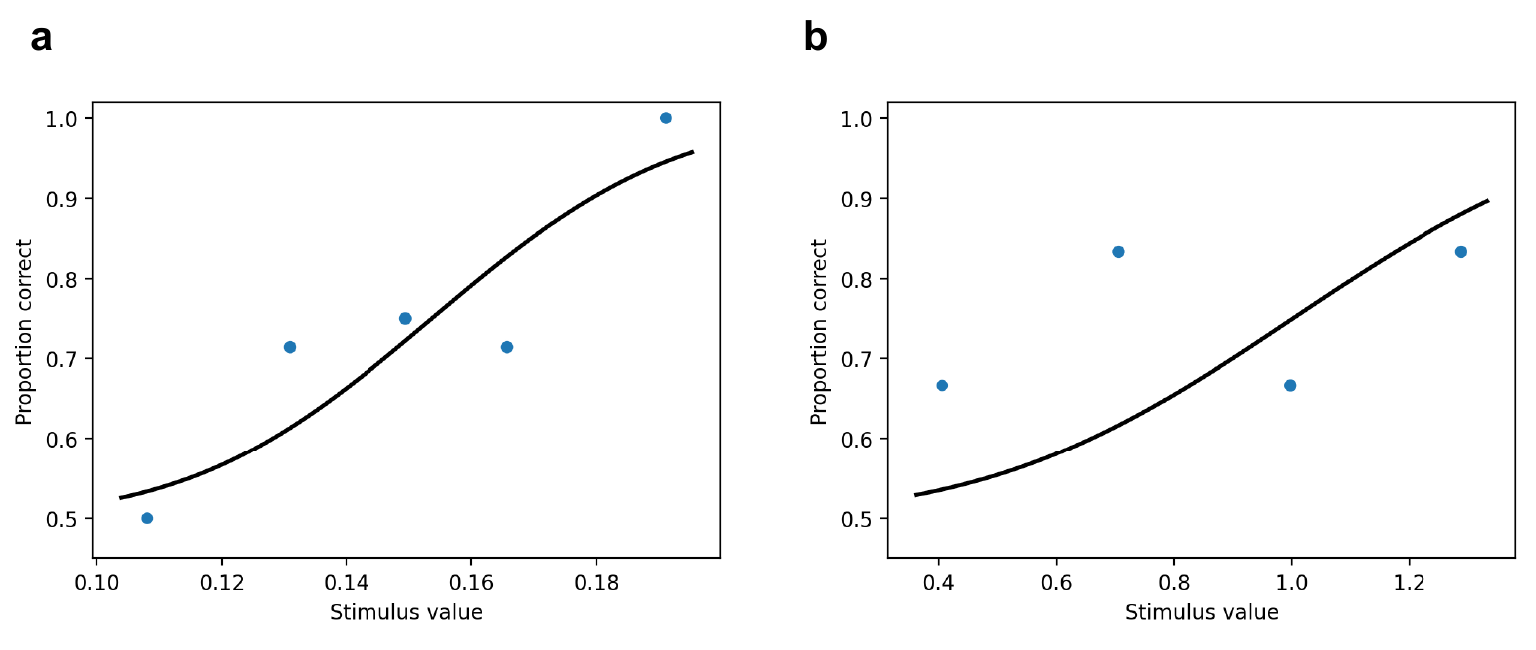}
    \caption{Example psychometric function fits from a single participant. Thresholds in the main analysis were estimated by fitting sigmoidal functions. Because some fits may not adequately capture the data (e.g., b), we additionally analyzed the data using reversal averages from the classical staircase procedure and obtained similar results (Tables~\ref{tab:overall_subjects}). We also confirmed that fit quality was comparable across layer conditions (RMSE: \texttt{conv1} = 0.11, \texttt{layer3} = 0.10, \texttt{avgpool} = 0.13).}
    \label{fig:psych_fit}
\end{figure}

\begin{table}[H]
\centering
\caption{Results of the two-way repeated-measures ANOVA for SSIM metrics. Reported $p$ values are Greenhouse--Geisser corrected.}
\label{tab:anova_results1}
\begin{tabular}{lccccc}
\hline
Source & df & $F$ & $p$ & $\eta_g^2$ \\
\hline
Layer & $(2, 14)$ & 696.67 & $< .001$ & 0.936 \\
Eccentricity & $(2, 14)$ & 30.91 & $< .001$ & 0.346 \\
Layer $\times$ Eccentricity & $(4, 28)$ & 7.00 & .016 & 0.218 \\
\hline
\end{tabular}
\end{table}

\begin{table}[H]
\centering
\caption{Results of the two-way repeated-measures ANOVA for LPIPS metrics. Reported $p$ values are Greenhouse--Geisser corrected.}
\label{tab:anova_results2}
\begin{tabular}{lccccc}
\hline
Source & df & $F$ & $p$ & $\eta_g^2$ \\
\hline
Layer & $(2, 14)$ & 826.47 & $< .001$ & 0.937 \\
Eccentricity & $(2, 14)$ & 38.25 & $< .001$ & 0.368 \\
Layer $\times$ Eccentricity & $(4, 28)$ & 5.30 & .044 & 0.183 \\
\hline
\end{tabular}
\end{table}

\begin{table}[H]
\centering
\caption{Results of the two-way repeated-measures ANOVA for DISTS metrics. Reported $p$ values are Greenhouse--Geisser corrected.}
\label{tab:anova_results3}
\begin{tabular}{lccccc}
\hline
Source & df & $F$ & $p$ & $\eta_g^2$ \\
\hline
Layer & $(2, 14)$ & 1871.61 & $< .001$ & 0.980 \\
Eccentricity & $(2, 14)$ & 23.16 & $< .001$ & 0.316 \\
Layer $\times$ Eccentricity & $(4, 28)$ & 8.47 & .015 & 0.248 \\
\hline
\end{tabular}
\end{table}

\end{document}